\documentclass{article}

\usepackage{PRIMEarxiv}

\usepackage[utf8]{inputenc} 
\usepackage[T1]{fontenc}    
\usepackage{hyperref}       
\usepackage{url}            
\usepackage{booktabs}       
\usepackage{amsfonts}       
\usepackage{nicefrac}       
\usepackage{microtype}      
\usepackage{lipsum}
\usepackage{fancyhdr}       
\usepackage{graphicx}       
\graphicspath{{media/}}     

\usepackage{graphicx}
\usepackage{subcaption}
\usepackage{threeparttable}
\usepackage{amssymb}
\usepackage{amsmath}
\usepackage{amsthm}
\usepackage{float}
\usepackage{mathrsfs}
\usepackage[ruled,vlined,linesnumbered]{algorithm2e}
\usepackage{orcidlink}

\theoremstyle{plain} 
\newtheorem{theorem}{Theorem}[section]  
\newtheorem{proposition}[theorem]{Proposition}

\title{On Procrustes Contamination in Machine Learning Applications of Geometric Morphometrics}

\author{
  Lloyd Austin Courtenay\orcidlink{0000-0002-4810-2001} \\[1em]
  PACEA UMR5199, CNRS \\
  Universit{\'e} de Bordeaux \\
  Pessac, France \\
  \texttt{ladc1995@gmail.com} \\
  \\
  Department d'Hist{\`o}ria i Hist{\`o}ria de l'Art \\
  Universitat Rovira i Virgili \\
  Tarragona, Spain
}

\begin{document}
\maketitle

\begin{abstract}
Geometric morphometrics (GMM) is widely used to quantify shape variation, more recently serving as input for machine learning (ML) analyses. Standard practice aligns all specimens via Generalized Procrustes Analysis (GPA) prior to splitting data into training and test sets, potentially introducing statistical dependence and contaminating downstream predictive models. Here, the effects of GPA-induced contamination are formally characterised using controlled 2D and 3D simulations across varying sample sizes, landmark densities, and allometric patterns. A novel realignment procedure is proposed, whereby test specimens are aligned to the training set prior to model fitting, eliminating cross-sample dependency. Simulations reveal a robust “diagonal” in sample-size vs. landmark-space, reflecting the scaling of RMSE under isotropic variation, with slopes analytically derived from the degrees of freedom in Procrustes tangent space. The importance of spatial autocorrelation among landmarks is further demonstrated using linear and convolutional regression models, highlighting performance degradation when landmark relationships are ignored. This work establishes the need for careful preprocessing in ML applications of GMM, provides practical guidelines for realignment, and clarifies fundamental statistical constraints inherent to Procrustes shape space.
\end{abstract}


\section{Introduction}
The modern landscape of scientific inquiry has long been undergoing a fundamental paradigm shift, transitioning from traditional hypothesis-driven methodologies to a data-centric, Artificially Intelligent accelerated model of discovery. Machine Learning (ML) has evolved through this, from a niche computational tool, to a foundational pillar of most modern research. However, this rapid adoption is not without complications. Although Artificial Intelligence (AI) has contributed to substantial increases in scientific productivity, concerns remain regarding the interpretability and appropriateness of these methods. As predictive performance and analytical output continue to accelerate, it is increasingly important to critically assess whether AI-based approaches (i) operate in ways consistent with their intended theoretical assumptions, (ii) represent the most suitable methodological choice for a given research question, and (iii) generate predictions whose apparent sophistication may obscure limited accuracy or robustness. Moreover, the diversity of available algorithms reflects the implications of the no free lunch theorem \cite{wolpert_no_free_lunch_1996, wolpert_no_free_lunch_1997}, which formalizes that no single algorithm performs optimally across all possible problem domains.

Here, the term Machine Learning (ML) is used as an umbrella designation encompassing a broad class of Computational Learning (CL) approaches, including the more complex subset commonly referred to as Deep Learning (DL), which is primarily based on Neural Network (NN) architectures. At its core, ML is grounded in the design of computational algorithms that integrate concepts from applied mathematics, linear algebra, calculus, and probability theory to infer decision boundaries from data \cite{mnist_nn_1998, bishop_pattern_recog_book_2006, lecun_dl_2015, goodfellow_dl_2016}. These decision boundaries enable models to generate predictions by identifying patterns and regularities within the input data. Model parameters are estimated through a process referred to as learning \cite{turing_machine_intelligence_1950, poole_computational_intelligence_1998, russel_ai_2003, legg_ai_2007, kaplan_siri_2018}, in which iterative optimization procedures progressively adjust internal representations to minimize predictive error \cite{sgd_2004, adam_2015}.

The recent surge in the adoption of ML-based methods closely parallels the trajectory described by the \textit{Gartner Hype Cycle}, a conceptual model characterizing the maturation of emerging technologies \cite{fenn_gartner_2008}. According to this framework, novel methodologies initially experience a phase of inflated expectations and widespread adoption, followed by a period of disillusionment as limitations and pitfalls become apparent, and ultimately progress toward a phase of stabilization as best practices and appropriate use cases are established. At present, evolutionary, palaeontological and archaeological sciences, can be considered to fall within the range of the emerging phase of inflated expectations. Despite its apparent accessibility, in many domains, the underlying mechanisms that condition and constrain ML-derived results remain insufficiently explored, as will be discussed here.

Importantly, many machine learning failures do not arise from model choice, but from preprocessing steps that implicitly violate the assumptions under which learning algorithms operate. In particular, most ML methods assume that training and test observations are conditionally independent given the data-generating process \cite{goodfellow_dl_2016}. When preprocessing procedures introduce statistical dependence across observations prior to model fitting, this assumption is broken, potentially leading to optimistic bias in model evaluation \cite{kapoor_reproducibility_crisis_ml_2023, tampu_leakage_ct_2022, bouke_data_leakage_2023, courtenay_deep_learning_taphonomy_2024}. Despite growing awareness of data leakage and contamination in applied ML, the role of geometric preprocessing operators in inducing such dependence has received little attention. In this study, Generalised Procrustes Analysis (GPA) is examined as a canonical example of a widely used alignment procedure that alters the geometry of the sample space itself, thereby coupling observations in ways that are invisible to downstream learning algorithms.

Geometric Morphometrics (GMM) provides a framework for the quantitative analysis of biological shape using sets of homologous landmarks represented as coordinates in Euclidean space. Because raw landmark configurations differ by arbitrary position, orientation, and scale, shape comparisons require an explicit alignment step prior to analysis \cite{bookstein_coords_1984, bookstein_book_1991, rohlf_morpho_spaces_1996, rohlf_shape_stats_1999, dryden_mardia_R_2016}. GPA fulfils this role by jointly superimposing all specimens in a sample and projecting them into a common coordinate space. As a result, GPA is not a specimen-wise transformation but a global operator whose outcome depends on the composition of the dataset being aligned.

The application of ML-based methods in GMM extends further back than is often appreciated. Early studies employing NNs on landmark data include \cite{dobigny_gmm_nn_2002} and \cite{baylac_gmm_nn_2003}, while \cite{manusco_efa_nn_1999} demonstrated the use of NNs on Elliptic Fourier coefficients for shape analysis and genotype identification. At a later stage, \cite{boxclaer_gmm_ml_2010} applied support vector machines (SVMs) to discriminate among species of shells. Subsequently, Lorenz et al. \cite{lorenz_gmm_nn_2015} and \cite{soda_gmm_nn_2017} employed NNs for classification tasks, highlighting the early growing adoption of ML techniques within GMM. 

Since these pioneering studies, subsequent work has focused on several directions. These include the development of automated landmarking techniques \cite{devine_deep_learning_landmarks_2020, bermejo_3d_skull_landmarks_2021, yeh_spine_landmarks_2021, kristiansen_husmorph_2025, kleisner_facedig_2025, rostamian_lower_limb_landmarks_2025, ma_deeplearning_curve_of_spee_2026}, simulation approaches to address small datasets \cite{courtenay_gan_2020, courtenay_pachycrocuta_preprint_2025}, and extensive investigations into classification based on morphological traits \cite{courtenay_cut_trampling_2020, courtenay_carnivore_tooth_class_2021, delbove_nn_cranial_sex_2022, courtenay_efa_cancer_2022, bonhomme_cnn_outline_2025, godinho_mandibular_sex_2026}.

Together, these developments demonstrate both the potential and challenges of applying ML to GMM, highlighting the importance of interpretability and methodological rigor. Machine learning offers several distinct advantages for systematic biologists. By capturing complex, multivariate patterns in shape data, ML can enhance classification accuracy and enable the automated detection of subtle morphological signals that may be overlooked by traditional methods. Such approaches facilitate the analysis of high-dimensional datasets, improve pattern recognition tasks, and perform in depth large scale comparisons of samples. Such approaches allow researchers to quantify shape variation with unprecedented precision. Furthermore, the ability of ML models to integrate heterogeneous data sources, including genetic, environmental or developmental covariates, opens new opportunities for investigating the drivers of morphological diversity and evolutionary patterns.

Despite these advantages, the application of ML to GMM data in particular raises important theoretical and methodological questions before its widespread use. In particular, it remains unclear to what extent landmark based shape data satisfies the assumptions underlying most ML models, and whether the available sample sizes are sufficient to support robust predictive inferences. These considerations underscore the need for a systematic reflection on the suitability of GMM as a source of data for the training of ML algorithms. This in turn will have an impact on the development of interpretability frameworks that can provide insight into the biological meaning of model predictions.

\section{Contamination in Machine Learning}

A central requirement of machine learning is that model evaluation be conducted on data that are statistically independent of the training process \cite{bishop_pattern_recog_book_2006, goodfellow_dl_2016}. Violations of this assumption, commonly referred to as \textit{data contamination} or \textit{data leakage}, occur when information from validation or test sets influences model training or preprocessing, leading to biased performance estimates and compromised generalisation \cite{kapoor_reproducibility_crisis_ml_2023, bouke_data_leakage_2023}.

In both supervised and unsupervised learning settings \cite{bishop_nn_book_1995, bishop_pattern_recog_book_2006, sutton_reinforcement_2015}, contamination most frequently arises during data preprocessing. Any transformation whose parameters are estimated using the full dataset—such as normalisation, centring, dimensionality reduction, or feature extraction—implicitly incorporates information from held-out samples if applied prior to dataset partitioning \cite{goodfellow_dl_2016, tampu_leakage_ct_2022}. As a result, downstream models may exploit artefactual dependencies that would not be available under genuine out-of-sample prediction.

Best practices therefore require strict separation between training, validation, and test datasets, with all data-dependent transformations estimated exclusively on the training set and subsequently applied to held-out data \cite{goodfellow_dl_2016, optimization_book_2019}. Failure to enforce this separation undermines the validity of performance metrics and obscures the true generalisation behaviour of the model \cite{modeling_generalization_2020}.

In the context of geometric morphometrics, Generalised Procrustes Analysis (GPA) constitutes a dataset-level transformation whose outcome for any given specimen depends explicitly on the composition of the entire sample. When GPA is applied prior to train–test splitting, the aligned coordinates of test specimens are influenced by training data, and vice versa. This violates the independence assumptions underlying standard ML evaluation protocols and represents a direct, but largely underformalised, source of data contamination in morphometric ML pipelines.

\section{Machine Learning in Geometric Morphometrics}

In landmark-based GMM, ML models can take two primary forms of input data. The first approach uses the superimposed landmark coordinates themselves to define the input variables $x$, while the second relies on principal components (PCs) derived from the landmark data to reduce dimensionality and capture major axes of shape variation.

Formally, we define a set of landmark configurations as matrices $X_i \in \mathbb{R}^{p \times k}$, where $p$ is the number of landmarks, $k$ the number of spatial dimensions, and $i = 1, \dots, n$ indexes individuals in the sample. When using landmark coordinates directly as input for ML algorithms, the data are typically preprocessed via Procrustes superimposition using the Generalised Procrustes Analysis (GPA) algorithm. GPA removes variation due to translation, rotation, and scale across specimens, leaving only the \textit{shape} information of interest. Variations of GPA however can be performed removing the scaling process, leaving only the \textit{form} information, however for the purposes of this study we fill focus only on \textit{shape}.

The most commonly used algorithm to perform GPA is that described by \cite{proc_extension_1990, dryden_mardia_R_2016}, where an iterative superimposition procedure is used to progressively improve the alignment of specimens. GPA, in this sense, considers minimizing the sum of squared norms of pairwise differences between shapes, $Q$, through;

\begin{equation}
    Q = \min \left( \frac{1}{n} \sum_{i = 1}^{n} \sum_{j=1}^{n} \lVert \left( \beta_{i} X_{i} \Gamma_{i} + \vec{1}_p \alpha_i \right) - \left( \beta_{j} X_{j} \Gamma_{j} + \vec{1}_p \alpha_j \right)\rVert^2 \right)
\end{equation}

\noindent where $\beta$ is a scalar parameter defined by the centroid size, $\Gamma \in \mathbb{R}^{k \times k}$ is an orthogonal rotation matrix, $\alpha \in \mathbb{R}^{1 \times k}$ is a translation vector, and $\vec{1}_p \in \mathbb{R}^{p \times 1}$ is a column vector of ones. GPA proceeds iteratively through three main steps. First, the landmarks of each specimen are centered to form preshapes. Second, each configuration is scaled and rotated to best match a reference, typically the mean shape. Third, these steps are repeated until $Q$ can no longer be reduced, yielding the optimally aligned landmark configurations.

Once superimposed, each matrix $X_i$ is then vectorized by stacking its $k$ coordinates for each landmark, producing an input matrix $X' \in \mathbb{R}^{n \times (kp)}$, where each row represents an individual and each column a single coordinate.

In the principal components approach, the covariance matrix of $X'$ is decomposed using Singular Value Decomposition (SVD):

\begin{equation}
    X' = U \cdot \Sigma \cdot V^{\top},
\end{equation}

\noindent where $U \in \mathbb{R}^{n \times n}$ contains the left singular vectors (scores), $\Sigma \in \mathbb{R}^{n \times kp}$ is the diagonal matrix of singular values, and $V \in \mathbb{R}^{kp \times kp}$ contains the right singular vectors (loadings). The resulting principal component scores in $U \cdot \Sigma$ can then be used as reduced-dimension inputs for ML models, capturing the major axes of shape variation while mitigating high-dimensionality and multicollinearity in the raw landmark data.

A major caveat when using principal components as inputs is that the PCs are typically computed from the covariance structure of all available landmark data, including specimens that may later be used for validation or testing. This means that information from the entire dataset contributes to the definition of the principal axes, introducing a form of data contamination prior to any model fitting. This issue has been highlighted by several authors \cite{calder_machine_learning_2022, courtenay_augmentationmc_X}, yet it is often underappreciated in practice.

Importantly, the standard Procrustes superimposition used to align landmarks can also introduce subtle sources of contamination. Because GPA aligns all specimens simultaneously, the position of each individual is influenced by all others in the dataset, including those intended for validation or testing. Consequently, even when models are trained on vectorized landmark coordinates rather than PCs, information from the full dataset is embedded in the aligned coordinates. This highlights that the preprocessing steps themselves — not just the downstream dimensionality reduction — can compromise the independence of training and testing sets, with potential consequences for model evaluation and generalisation.

\section{Simulation Methods and Analyses}

This section first establishes a conceptual framework for diagnosing contamination, providing a framework under which the rest of the study will follow. This will be followed by a mathematical proof of Procrustes contamination, before evaluating the effects this may have on ML applications.

\subsection{Quantifying Contamination}

Data contamination is most effectively detected through its impact on model performance and generalisation. In the present study, we focus on supervised regression analyses as a means of identifying, quantifying, and mitigating contamination in ML workflows applied to GMM.

Among supervised learning paradigms, regression provides a particularly suitable framework for studying contamination effects. In regression, the response variable $y$ is continuous and can be controlled to vary smoothly, allowing prediction errors to be interpreted directly and quantitatively. This makes regression especially well suited for detecting subtle distortions in the relationship between shape variables and biological parameters.

All experiments in this study are framed as allometric regression analyses, in which multivariate shape variables are used to predict a single continuous outcome: centroid size. Although allometric regression is not a typical objective in many ML-based GMM applications, it offers a controlled and biologically interpretable setting in which to evaluate how algorithms learn the mapping $f(x)=y$. Moreover, centroid size is a fundamental quantity in GMM, making it a natural benchmark for assessing downstream learning behaviour.

Regression models are particularly sensitive to loss or distortion of signal, as even small reductions in informative variation directly affect predictive accuracy. For this reason, regression serves as a stringent test of how preprocessing steps, such as Procrustes superimposition, influence learning outcomes. In contrast, classification approaches discretize continuous variation into categorical labels. As long as class boundaries remain separable, moderate degradation of underlying biological signal may go undetected, potentially masking contamination effects and inflating apparent model performance. From this perspective, classification can obscure biologically meaningful information loss, whereas regression exposes it directly.

One way in which contamination manifests in regression analyses is through discrepancies in model error across datasets. When a model is influenced by information from the validation or test sets during training, predictive performance can appear artificially high. This inflation of accuracy reflects a failure of the model to generalise beyond the data it has effectively already encountered.

\begin{figure}[tbp]
    \centering
    \captionsetup{justification=centering}
    \includegraphics[width=0.7\textwidth]{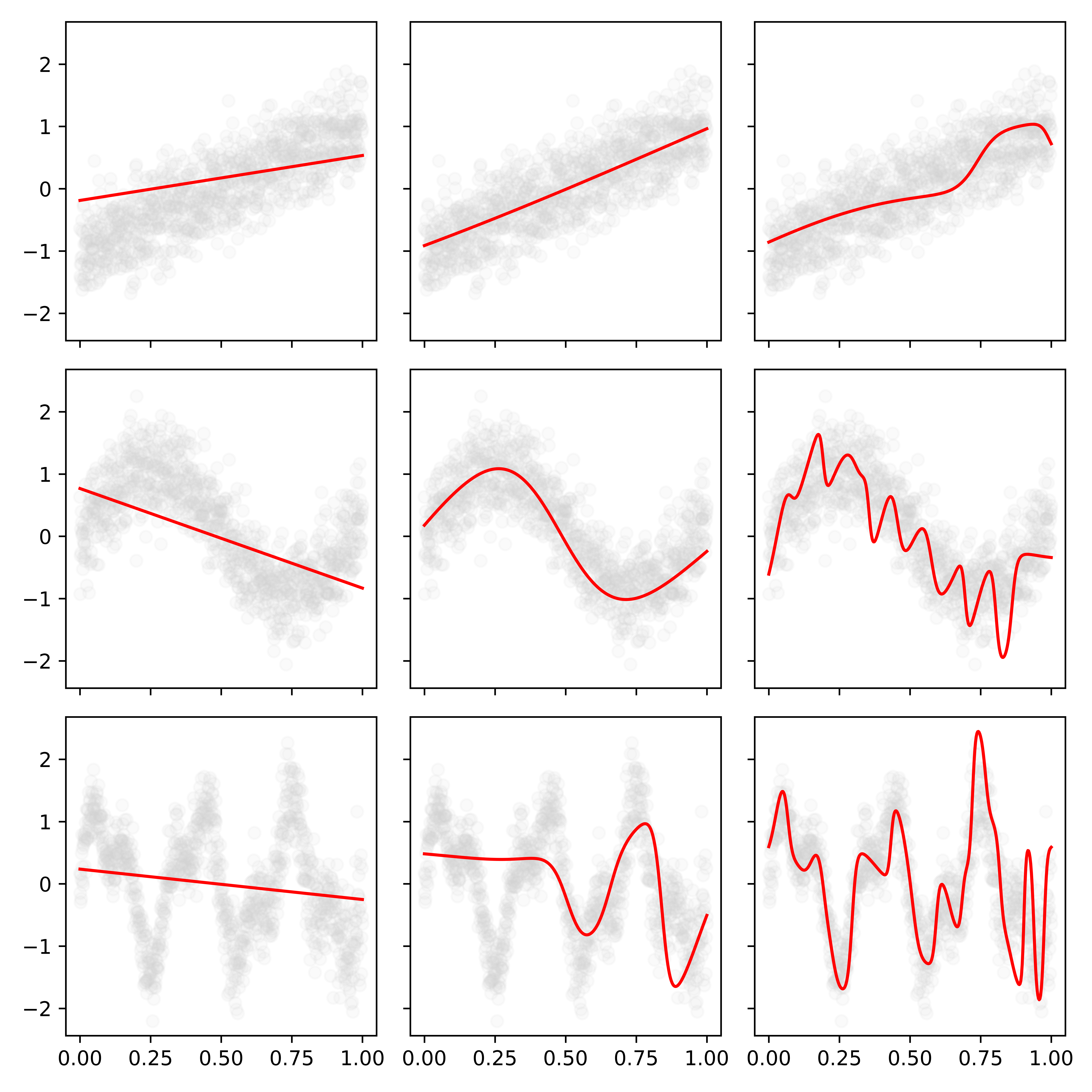}
    \caption{Illustrative examples of underfitting (left), near-optimal fitting (center), and overfitting (right) in regression on synthetic data. Top row: linear relationship $y = 2x - 1 + \mathcal{N}(0, 0.4^2)$. Middle row: moderately nonlinear relationship $y = \sin(2\pi x) + 0.2x + \mathcal{N}(0, 0.4^2)$. Bottom row: highly nonlinear relationship $y = \sin(6\pi x) + \tfrac{1}{2}\sin(14\pi x) + \mathcal{N}(0, [0.25(1+x)]^2)$, exhibiting heteroscedastic noise.}
    \label{fig:examples_of_overfitting}
\end{figure}

This phenomenon is commonly referred to as \textit{overfitting}. Overfitting is a common symptom of contamination and occurs when a model captures noise, idiosyncratic structure, or dataset-specific artifacts rather than underlying, generalisable patterns \cite{lever_overfitting_2016, modeling_generalization_2020} (Fig. \ref{fig:examples_of_overfitting}). As a result, the model exhibits low error on the data used for training or model selection, but performs poorly on truly independent data. Data contamination exacerbates overfitting by reducing the effective independence between training, validation, and testing sets, leading to overconfident estimates of predictive ability.

In the present study, overfitting is used as a diagnostic indicator of potential data contamination. To quantify model performance, we employ the Root Mean Squared Error (RMSE), a standard metric for evaluating predictive accuracy in regression analyses.

Let $y_j$ denote the true value of a continuous response variable for observation $j = 1, \dots, n$, and let $\hat{y}_j = f(x_j)$ denote the corresponding prediction produced by the trained model $f(\cdot)$. The prediction error for a single observation is given by the residual $(\hat{y}_j - y_j)$. In regression settings, these residuals are commonly aggregated through a squared-error loss function.

The overall prediction error is quantified using the Mean Squared Error (MSE):

\begin{equation}
    \mathrm{MSE} = \frac{1}{n} \sum_{j=1}^{n} (\hat{y}_j - y_j)^2,
\end{equation}

\noindent and its square root, the Root Mean Squared Error (RMSE),

\begin{equation}
    \mathrm{RMSE} = \sqrt{\mathrm{MSE}},
\end{equation}

\noindent which returns the error metric to the original scale of the response variable, facilitating biological interpretation.

Overfitting is identified by comparing RMSE values across training, validation, and test datasets. A model that exhibits low RMSE on the training data but substantially higher RMSE on independent test data has failed to generalise. In the presence of data contamination, however, this discrepancy may be reduced or eliminated, leading to artificially low test RMSE values and an overestimation of predictive performance.

\subsection{Simulation of Data}

To assess the relationship between predictor variables $x$ and response variables $y$ in a controlled and standardised setting, all analyses in this study were conducted using simulated shape data.

Base shapes $\bar{X}$ were defined as regular two-dimensional configurations consisting of $p$ landmarks arranged equiangularly around the unit circle. These base shapes were generated by iterative rotation of an initial landmark around the centroid, producing simple geometries with known structure.

For each simulation, a sample of $n$ individual shapes were generated by perturbing each landmark coordinate with isotropic Gaussian noise, specifically;

\begin{equation}
    x_{i, j} \sim \mathcal{N} \left( \mu_{x, j} , \sigma^2 \right),\quad  y_{i, j} \sim \mathcal{N} \left( \mu_{y, j} , \sigma^2 \right)
\end{equation}

\noindent for landmark $j$ of individual $i$, where $\left( \mu_{x, j}, \mu_{y, j}\right)$ denote the coordinates of the corresponding landmark in the base shape, and $\sigma$ controls within-landmark variance.

Systematic shape variation was introduced through the application of a simple shear transformation. Each individual shape was sheared along a single axis according to $x' = x + \epsilon_i y$ with the orthogonal coordinate left unchanged. The shear parameter $\epsilon_i$ varied across individuals and was generated as a deterministic sequence spanning a predefined range, with additive Gaussian noise, introducing smooth but imperfectly ordered shape variation across the sample.

To model allometric trends detectable by regression algorithms, individual shapes were additionally scaled by a size factor, $s_i$, defined as $s_i = z_i^{\rho} + \delta_s$, where $z_i$ is an auxiliary ordering variable and $\delta_s \sim \mathcal{N}\left(0, \sigma_s ^2\right)$. This non-linear scaling, whose degree is conditioned by $\rho$, produced a broad distribution of centroid sizes while preserving a known functional relationship between size and shape.

Initial experiments were run setting $\sigma^2 = 0.5$, $\epsilon \in [-0.75, 0.75]$, and $\rho = 4$. Across simulation experiments, these key parameters were systematically varied to assess their influence on downstream analyses. Changes therefor considered both lower ($\epsilon \in [-0.1, 0.1]$) and higher degrees of shear ($\epsilon \in [-1.4, 1.4]$), different polynomials for size ($\rho = 2$ or $\rho = 5$), and finally the amount of noise used to perturb landmark coordiantes $\sigma = 0.05$ or $\sigma = 1$. 

In regression analyses, linear models were therefore trained to predict the size parameter $s$ given the vectorised shape coordinates $x$. For these models, Ordinary Least Squares linear regression models were deliberately chosen to assess the statistical effects of GPA without conflating them with regularisation, hyperparameter tuning, or optimiser behaviour.

\subsection{Detecting Procrustes Contamination}

To evaluate the practical effects of this contamination on downstream learning tasks, two GPA workflows were compared. In the standard workflow, the full dataset is superimposed prior to splitting into training and test sets using Algorithm \ref{alg:gpa}. In the controlled workflow, only the training set is used to compute the GPA reference shape; the test set is then aligned to this fixed reference without iterative processing, as expressed in Algorithm \ref{alg:train_test_gpa}. In the second scenario, all geometric references are estimated exclusively from the training set, and test shapes are aligned without updating the reference configuration. Regression models are subsequently trained on the vectorized coordinates of the training set and evaluated on the independent test set. This design isolates the influence of GPA on model performance, allowing direct quantification of contamination through differences in RMSE between the two workflows, independently of model choice or optimisation strategy.

The null hypothesis is that there is no difference in RMSE between models trained under the standard and controlled GPA workflows. The alternative hypothesis is that the standard (contaminated) workflow yields artificially lower RMSE values due to information leakage. It is important to highlight that in all cases algorithms were trained and evaluated on the same individuals. A 70:30 train:test split was employed for all simulations. From this perspective, the only reason that algorithms would produce differences in RMSE values is due to contamination.

\begin{algorithm}[H]
    \caption{Generalized Procrustes Alignment}
    \label{alg:gpa}
    \KwIn{
        \\ 
        Landmark array $X \in \mathbb{R}^{p \times k \times n}$; \\
        Robust flag $\in \{\text{TRUE}, \text{FALSE}\}$; \\
        Scale flag $\in \{\text{TRUE}, \text{FALSE}\}$
    }
    \KwOut{
        \\
        Aligned configurations $X^\star$; \\
        Central configuration $\bar{X}$
    }
    
    \For{$i \leftarrow 1$ \KwTo $n$}{
        Translate $X_i$ to the origin using mean (or median if robust)
    }
    
    Iteratively rotate and update the reference configuration to minimize summed Procrustes distances using optimal orthogonal Procrustes rotation until convergence
    
    \If{Scale}{
        Normalize each configuration by centroid size and repeat rotation until convergence
    }
    
    Compute the central configuration $\bar{X}$ as the mean or coordinate-wise median
    
    \Return{
        $X^\star$, $\bar{X}$
    }
\end{algorithm}

\begin{algorithm}[H]
    \caption{Train–Test Alignment of Procrustes Coordinates}
    \label{alg:train_test_gpa}
    \KwIn{
        \\
        Training landmark array $X_{\text{train}} \in \mathbb{R}^{p \times k \times n_{\text{train}}}$; \\
        Test landmark array $X_{\text{test}} \in \mathbb{R}^{p \times k \times n_{\text{test}}}$; \\
        Robust flag $\in \{\text{TRUE}, \text{FALSE}\}$; \\
        Scale flag $\in \{\text{TRUE}, \text{FALSE}\}$
    }
    \KwOut{
        \\
        Aligned training shapes $X_{\text{train}}^\star$; \\
        Aligned test shapes $X_{\text{test}}^\star$; \\
    }
    
    \textbf{Training alignment:}
    
    Apply Algorithm~\ref{alg:gpa} to $X_{\text{train}}$ to obtain
    aligned configurations $X_{\text{train}}^\star$ and
    central configuration $\bar{X}$
    
    %
    %
    
    \textbf{Test alignment:}
    
    \For{$j \leftarrow 1$ \KwTo $n_{\text{test}}$}{
    
        Translate $X_j$ to the origin using the training centroid definition
        
        \If{Scale}{
            Normalize $X_j$ by its centroid size
        }
        
        Rotate $X_j$ onto $\bar{X}$ using the optimal Kendall rotation
        
        Store the aligned test configuration
        
    }

    No information from $X_{\text{test}}$ is used to update $\bar{X}$ or any other geometric statistic.
    
    
    \Return{
        $X_{\text{train}}^\star$,
        $X_{\text{test}}^\star$
    }
\end{algorithm}

\subsection{Assessing the Role of Spatial Autocorrelation in Shape-Size Regression}

Landmark configurations exhibit intrinsic spatial autocorrelation, as neighbouring landmarks are constrained by geometric continuity and biological structure. When landmark coordinates are vectorized for use in standard regression or machine learning models, this spatial information is discarded, and all predictors are treated as independent.

To test this hypothesis and isolate the effect of spatial structure in the learning architecture, two neural network regression models were implemented. In both cases, the objective was to predict centroid size from Procrustes-aligned landmark configurations using a mean squared error loss.

In the baseline condition, landmark coordinates were vectorized into a single feature vector and used as input to a linear regression neural network consisting of a single fully connected output layer. This model is functionally equivalent to ordinary least squares regression but trained using gradient-based optimisation.

In the structured condition, landmark coordinates were provided to the model in their original two-dimensional arrangement, preserving landmark ordering and coordinate pairing. A convolutional layer with a kernel spanning the full landmark configuration was applied prior to regression, enforcing a structured linear transformation that respects spatial organisation in the input. The convolutional output was subsequently flattened and mapped to a scalar prediction.

Both models were trained using identical optimisation settings (Adam optimiser, mean squared error loss), for 100 epochs with a batch size of 63. No regularisation or hyperparameter tuning was performed, ensuring that performance differences reflect architectural assumptions rather than optimisation choices.

\section{Results}

\subsection{Numeric Instability of GPA Under Sample Perturbation}

To quantify the dependence of GPA on sample composition, landmark configurations generated from a fixed generative process were superimposed under systematically perturbed datasets. GPA was applied to (i) the full dataset and (ii) a reduced dataset obtained by removing a single individual per iteration. Procrustes aligned coordinates for specimens common to both datasets were then compared.

\begin{figure}[tbp]
    \centering
    \captionsetup{justification=centering}
    \includegraphics[width=0.7\textwidth]{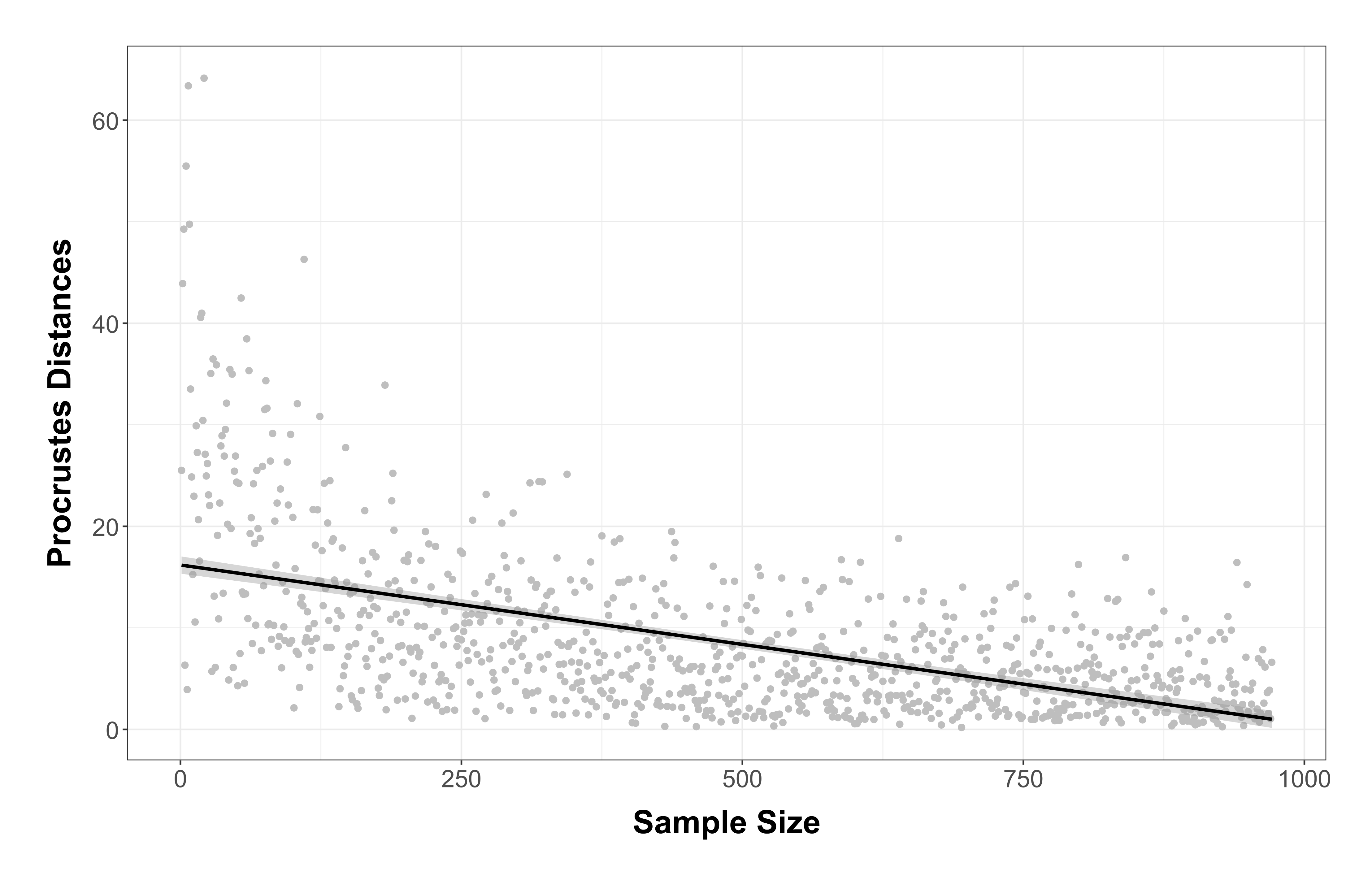}
    \caption{Bootstrapped ($\times$1000) Procrustes distances between Procrustes-aligned coordinates of specimens common to both datasets when GPA is performed on the full sample versus a reduced sample obtained by removing a single individual. Procrustes distances increase as sample size decreases, demonstrating that Procrustes-aligned coordinates are sensitive to sample composition.}
    \label{fig:proc_d_sample_size}
\end{figure}

If GPA operates independently on individual specimens, our null hypothesis is that aligned coordinates would be invariant to the inclusion or exclusion of other individuals. This invariance is not observed. Procrustes distances between identical specimens aligned under the two conditions increased as a function of decreasing sample size (Fig. \ref{fig:proc_d_sample_size}), indicating that the estimated coordinates of a specimen depend on the composition of the dataset used for superimposition.

While expected, the magnitude of differences are surprising especially when the number of individuals removed is small.

This result establishes that performing GPA prior to training-test partitioning introduces statistical dependence between specimens, providing a direct mechanism for data contamination in downstream analyses.

\subsection{Effects of Procrustes Contamination on Regression Performance}

The impact of GPA-induced contamination on predictive performance was assessed by comparing regression models trained under contaminated and uncontaminated alignment protocols. Differences in RMSE between the two approaches were computed across repeated simulations.

\begin{figure}[tbp]
    \centering
    \captionsetup{justification=centering}
    \includegraphics[width=1\textwidth]{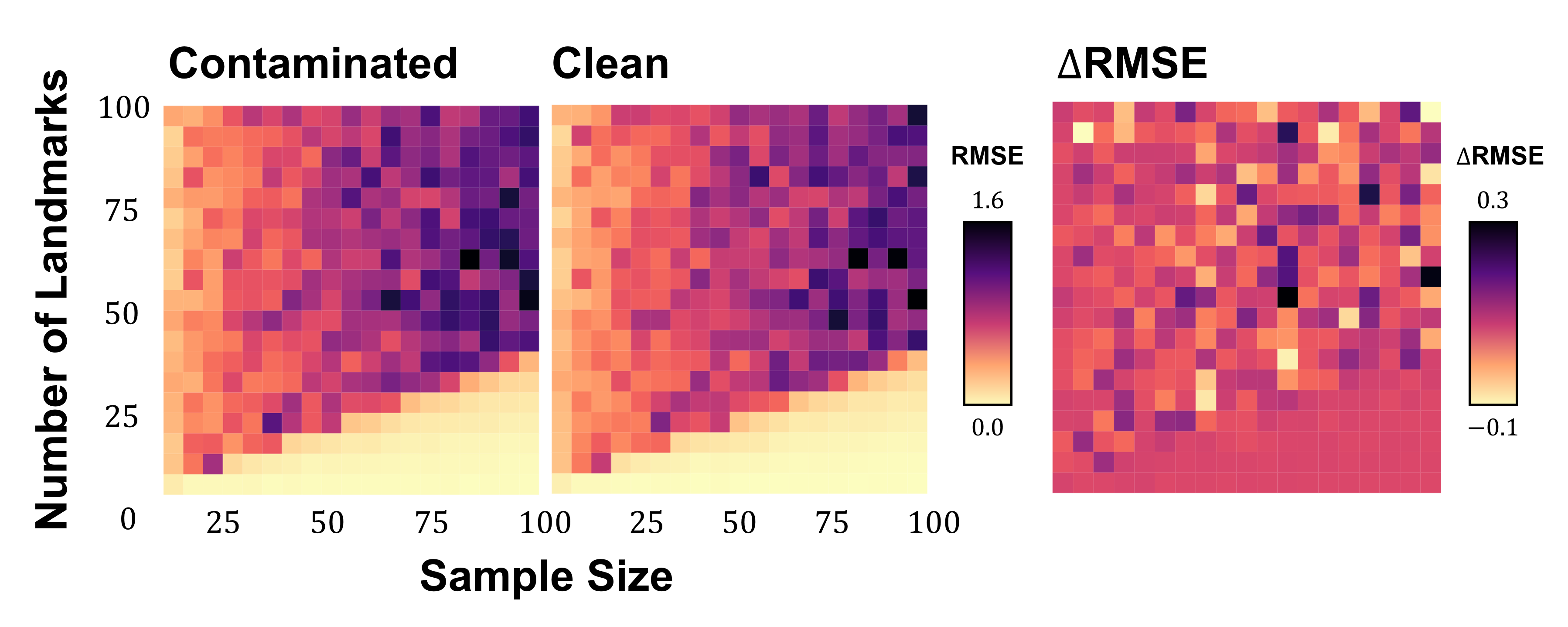}
    \caption{$RMSE$ and $\Delta RMSE$ values comparing the two pipelines; the contaminated GPA approach where GPA is performed prior to the train-test split, and the clean approach where GPA is performed on the training data, and test data is later projected onto the already superimposed data.}
    \label{fig:delta_rmse}
\end{figure}

Across all conditions, the mean difference in RMSE was -0.0117 with a 95\% bootstrap interval of [-0.053, 0.057] (Fig. \ref{fig:delta_rmse}). While this distribution spans zero, a systematic shift toward negative values was observed, indicating a tendency for models trained under contaminated alignment to perform better, despite being trained on the same individuals. This is conceptual proof of slight overfitting.

The spatial distribution of $\Delta$RMSE values across the sample-size-landmark grid reveals an interesting structured pattern. Regions of comparatively low instability emerge at high sample sizes and low landmark counts, while RMSE differences increase markedly as the ratio of landmarks to individuals increased. This pattern indicates that contamination effects are not uniform but depend strongly on the dimensionality-sample size balance.

\subsection{Sensitivity Analyses}

\begin{figure}[tbp]
    \centering
    \captionsetup{justification=centering}
    \includegraphics[width=1\textwidth]{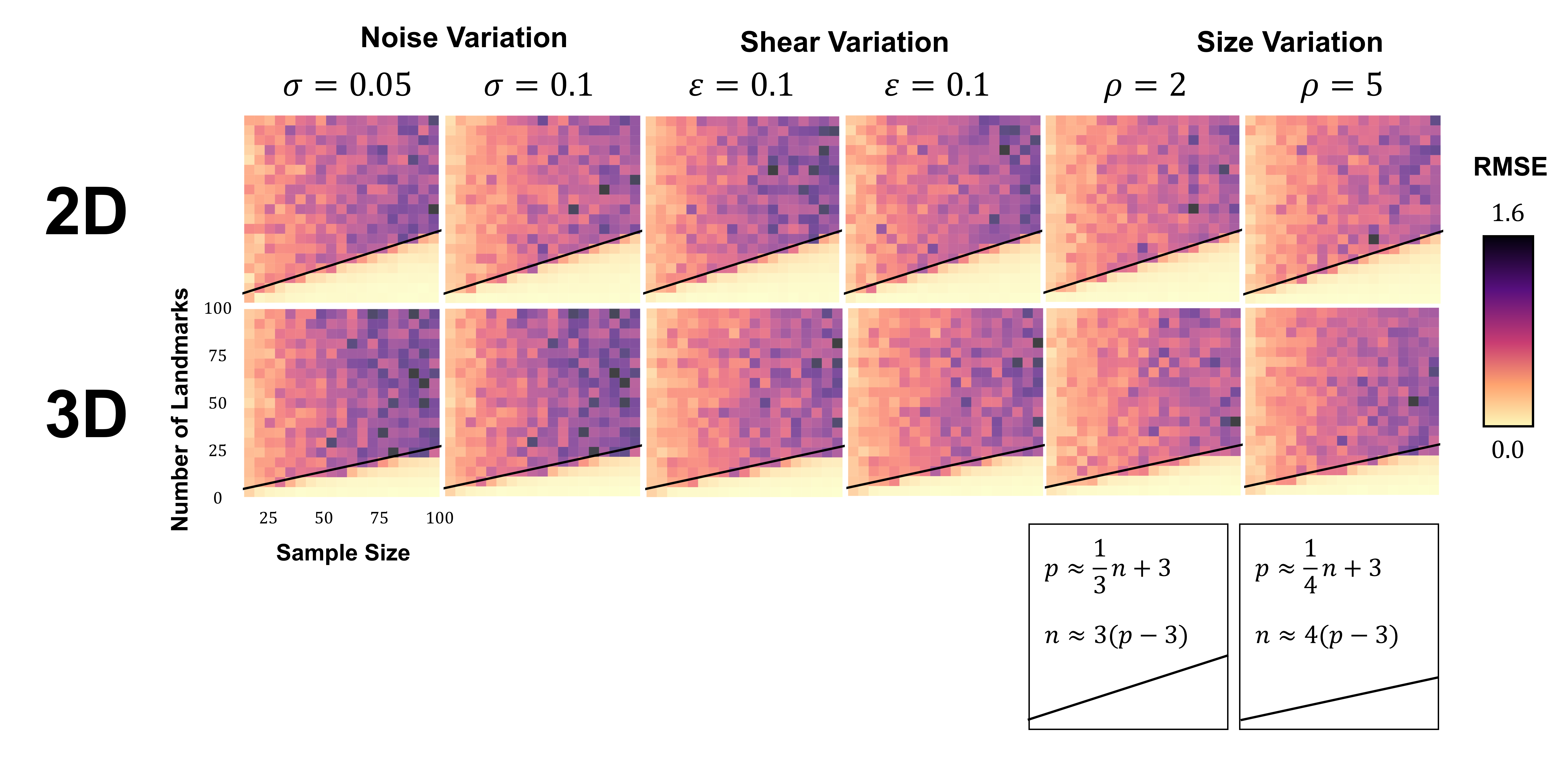}
    \caption{RMSE values comparing sample size with the number of landmarks under different experimental constraints in both 2D and 3D}
    \label{fig:sensitivity_tests_2d_3d}
\end{figure}

To assess the robustness of the observed instability pattern, simulations were repeated under varying magnitudes of shear, noise and allometric scaling (Fig. \ref{fig:sensitivity_tests_2d_3d}). Across all conditions tested, the qualitative structure of the RMSE surface grid remains stable.

In two dimensions, the boundary separating stable from unstable regions is well approximated by a linear relationship of the form;

$$
p \approx \frac{1}{3}n +3
$$

\noindent This relationship persists across all parameter variations, indicating that the observed diagonal structure reflects an intrinsic properly of the alignment-learning pipeline rather than a feature of specific generative assumptions.

Extending this idea to three dimensions, the corresponding boundary exhibits a shallower slope, consistent with an expected scaling near

$$
p \approx \frac{1}{4}n + 3
$$

\noindent (See Appendix \ref{ap:Appendix_A_PCA_Proc_Space} for a mathematical derivation of the expected diagonal structure). Empirically, the fitted slope, however, was approximately 0.22, reflecting finite-sample effects while preserving the predicted dimensional dependence.

\subsection{Impact of Spatial Arrangement on Algorithm Performance}

The effect of spatial structure in landmark data was evaluated by comparing regression models that either ignored or explicitly exploited landmark adjacency. When landmarks were treated as independent features using a linear regression model, prediction error increased substantially.

\begin{figure}[tbp]
    \centering
    \captionsetup{justification=centering}
    \includegraphics[width=0.5\textwidth]{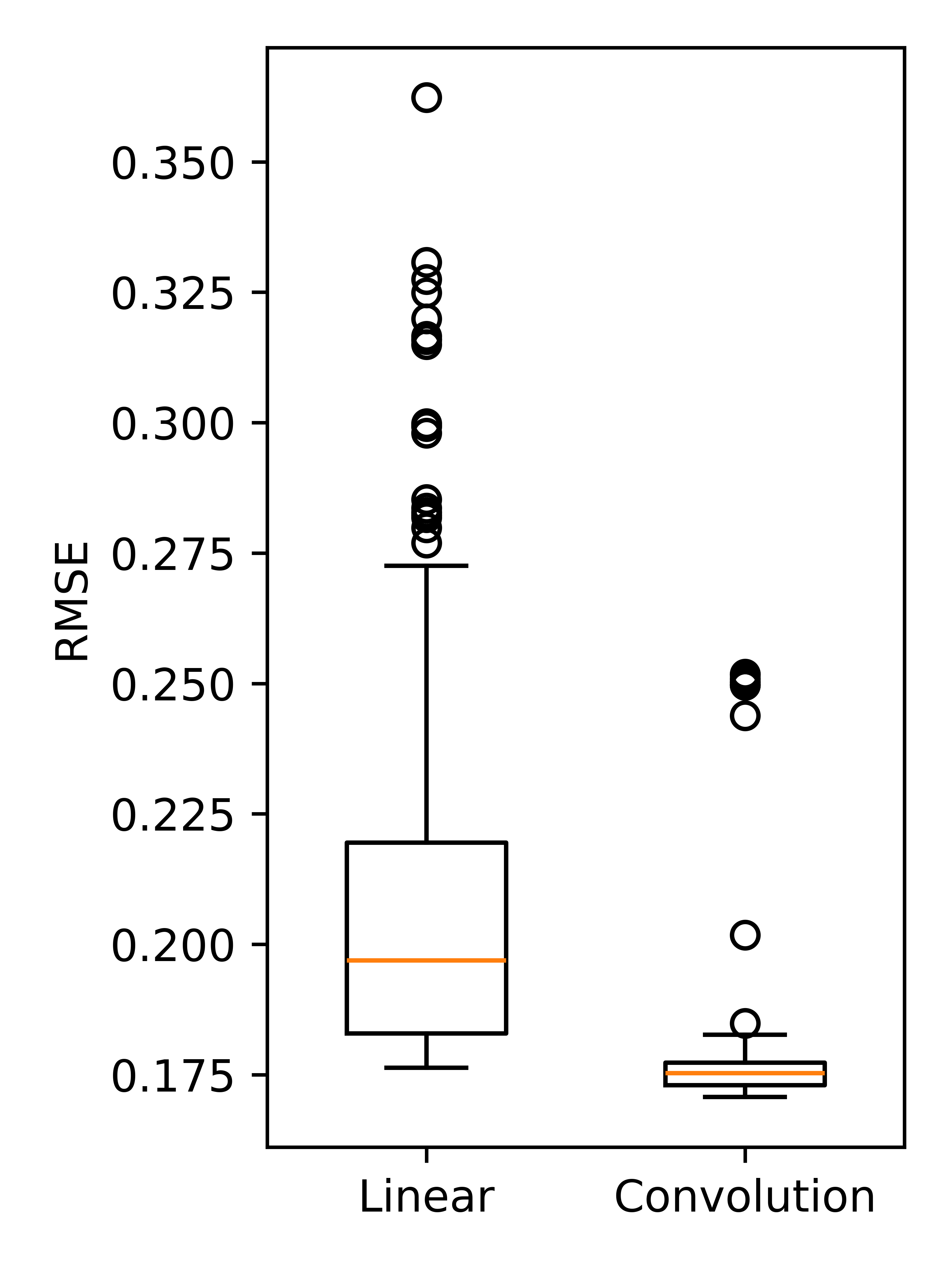}
    \caption{A boxplot comparing the RMSE values of 300 training simulations of basic linear models on landmark data and a model that is spatially aware of landmark coordinates.}
    \label{fig:linear_vs_conv_landmarks}
\end{figure}

Throughout simulations, linear models yield a mean RMSE of 0.2076 with a 95\% interval of [0.177, 0.316] (Fig. \ref{fig:linear_vs_conv_landmarks}). Models explicitly designed to exploit and preserve spatial autocorrelation between landmarks, however, achieve a much lower mean of 0.1778 $\in$ [0.1715, 0.2497]. This reduction indicates that preserving spatial relationships among landmarks materially improves predictive performance.

These results demonstrate that treating landmark coordinates as independent variables discards biologically, geometrically and spatially meaningful structure, leading to inflated error even under identical alignment and sampling conditions.

\section{Discussion}

ML applications in landmark based studies of the evolutionary sciences are beginning to receive more and more attention. Nevertheless, a systematic evaluation is needed of how the pre-processing of this specific type of data affects downstream analyses.

Simulations reveal a consistent relationship between sample size and landmark count in Procrustes-aligned shape data. In 2D, the slope of this diagonal approaches $1/3$, while in 3D it approaches $1/4$. While prior work has considered variable-to-individual ratios in analyses such as Canonical Variates Analysis and Between-Group PCA \cite{bookstein_pca_pathologies_2017, bookstein_bgpca_pathologies_2019, cardini_bgpca_2019, cardini_cv_bgpca_2020, rohlf_bgpca_cva_2021, courtenay_ordination_2022}, the present study demonstrates that these considerations extend to simpler tasks, including allometry estimation. These slopes emerge naturally from the geometry of Kendall's shape space and the constraints imposed by GPA, including the removal of translation, rotation, and scale (Appendix \ref{ap:Appendix_A_PCA_Proc_Space}). Simulations show these slopes to be robust across variations in experimental conditions, suggesting they reflect a fundamental property of the alignment process rather than idiosyncrasies of particular datasets.

The experiments further underscore that GPA introduces statistical dependence across specimens. This echoes hypotheses raised by recent authors \cite{medialdea_gmm_nutrition_out_of_sample_2026}, yet formalises this theory using simulation based approaches. First, here it is shown that removing or adding a single specimen systematically alters aligned coordinates of other shapes, establishing that pre-processing prior to train/test partitioning can introduce data contamination. This effect has practical consequences: naive GPA across combined training and test sets pulls test shapes toward the training mean, artificially reducing RMSE and inflating apparent predictive performance. By formalizing and quantifying this effect, and providing expected slope and intercept values for the sample-to-landmark relationship, this study establishes a framework to anticipate when contamination may bias model evaluation.

While the present study has used simulations to assess the reliability of GPA based workflows on GPA, the experiments in essence are rooted in the fundamental nature of simply testing for allometry. Therefore, it is important to point out that previous comments on the stability of the covariance matrix is something that branches out far beyond ordination practices in our field, but also must be considered for many other statistical tests, such as Procrustes ANOVA, shape-size testing, and potentially others.

Spatial structure among landmarks also contributes meaningfully to predictive performance. Models that exploit local correlations outperform those treating coordinates independently, confirming that landmark covariation is informative. These results formalize prior hypotheses about PCA-based pre-processing: while PCA can reduce dimensionality and mitigate overfitting, its orthogonality constraint may discard biologically relevant signal beyond the first components \cite{gould_pca_1967, bookstein_pca_pathologies_2017, bookstein_factor_coordinates_2017, courtenay_graphgmm_2024}.

In \cite{courtenay_graphgmm_2024} we explored this idea further by introducing the idea of using Graph convolutions to learn structure from landmark coordinates prior to performing PCA. This mathematical proposal stems directly from ML research in geometric learning, yet without adding learnable parameters to optimise landmark embeddings \cite{bronstein_geometric_deep_learning_2010, bronstein_geometric_deep_learning_2017, kipf_welling_gcnn_2017, leskovec_gnn_2019}. An optimal approach to ML applications in GMM should therefore explore Graph-based convolutions in more detail, potentially providing a powerful means to train models on this type of data. 

An important question raised by these results concerns the extent to which contamination effects observed under isotropic assumptions translate to real biological scenarios. Although Figure \ref{fig:delta_rmse} shows that RMSE improvements arising from contamination are often small, marginal, or unstable, this does not negate their relevance for model evaluation.

The simulations presented here deliberately adopt isotropic variation as a controlled baseline, allowing the geometric consequences of GPA and sample-to-landmark ratios to be isolated. Real landmark datasets, however, are rarely isotropic; biological shapes exhibit strong anisotropies, modularity, and spatially structured covaration. These departures from isotropy may either dampen or exacerbate contamination effects depending on how variance is distributed across landmarks and dimensions. Crucially, the present study is not intended to estimate the magnitude of contamination expected in any particular system, but instead to demonstrate that the pathology exists as a necessary consequence of alignment in finite samples. Even if the effect proves negligible in some real-world cases, its existence cannot be ignored \textit{a priori}, as it established a lower bound on bias that may interact unpredictably with biological structure, sampling design, and model architecture.

Finally, from a pattern recognition perspective, this contamination effect is particularly problematic because it operates orthogonally to model choice or learning capacity. Even simple distance-based or linear models exhibit artificially improved performance when test shapes are included in the Generalised Procrustes superimposition, as the geometry of the space itself is altered prior to learning. In this sense, the apparent success of a classifier or regressor may reflect properties of the alignment procedure rather than genuine structure in the data. This distinction is critical for pattern recognition tasks, where performance metrics are often interpreted as evidence of biologically meaningful signal. The present results therefore emphasise that, in landmark-based settings, preprocessing is not a neutral step but an integral component of the recognition pipeline, capable of inducing systematic bias that masquerades as predictive power if not explicitly controlled.

\section{Conclusions}

Careful attention to train/test alignment is essential in geometric morphometric analyses using machine learning. Diagonal patterns in RMSE plots, formalized through simulations and theoretical derivations, provide a rule-of-thumb to anticipate contamination effects. Combining proper test-set realignment with consideration of sample size, landmark density, and spatial covariation enables more reliable and interpretable predictive models. These guidelines serve as a practical foundation for the broader application of ML methods in morphometric research.

\section*{Appendix: Expected PCA Variance Accumulation in Procrustes Shape Space} \label{ap:Appendix_A_PCA_Proc_Space}

While the primary focus of this study is on superimposition, alignment, and their consequences for machine learning applications, the expected distribution of variance across shape-space dimensions provides a useful null reference for interpreting the slopes observed in the simulation studies. This appendix formalizes that expectation using a PCA-based argument in Procrustes tangent space.

\subsection*{Setup}

Let $p$ landmarks be observed in $k$-dimensional Euclidean space and aligned using Generalized Procrustes Analysis (GPA), removing translation, rotation, and scale. Let $X \in \mathbb{R}^{n \times q}$ denote the matrix of Procrustes-aligned configurations expressed in the tangent space at the mean shape, where the dimensionality of the tangent space is

\begin{equation}
    q = kp - k - \frac{k(k-1)}{2} - 1.
\end{equation}

\noindent Assume that the rows of $X$ are independently drawn from a centered, isotropic distribution in this tangent space with finite second moments. PCA is applied to $X$, yielding eigenvalues $\lambda_1 \geq \lambda_2 \geq \dots \geq \lambda_q$ of the sample covariance matrix;

\begin{equation}
    S = \frac{1}{n-1} X^\top X.
\end{equation}

\noindent Define the cumulative proportion of variance explained by the first $m$ principal components as 

\begin{equation}
    V(m) = \frac{\sum_{i=1}^{m} \lambda_i}{\sum_{i=1}^{q} \lambda_i}.
\end{equation}

For large numbers of landmarks $p$, the tangent-space dimension $q = kp - k - \frac{k(k-1)}{2} - 1$ is dominated by the term $kp$. Including the constant corrections from translation, rotation, and scale removal, the effective dimension per landmark approaches $k +1$, so that asymptotically, $q \sim (k + 1)p$. Here $\sim$ denotes asymptotic equivalence, meaning the leading term grows linearly with $p$, and constant subtractions from translation, rotation, and scale removal become negligible for large $p$

\subsection*{Proposition}

\begin{proposition}
    Under the assumptions above, as $n \to \infty$ and $p \to \infty$,
    
    \begin{equation}
        \mathbb{E}[V(m)] \rightarrow \frac{m}{q}.
    \end{equation}
    
    \noindent If the number of retained components scales linearly with landmark count such that $m = \alpha p$ for some $\alpha \in (0,1)$, then
    
    \begin{equation}
        \mathbb{E}[V(\alpha p)] \rightarrow \frac{\alpha}{k+1}.
    \end{equation}
    
    \noindent In particular, the expected slope is $1/3$ in two dimensions and $1/4$ in three dimensions.
    
    To clarify, $\alpha$ represents the proportionality constant linking the number of retained PCs to the number of landmarks, and $\alpha p$ is thus the number of PCs expressed as a linear function of landmark count, used to study asymptotic behaviour as $p \rightarrow \infty$
    
\end{proposition}

\subsection*{Proof}

\noindent\textit{Step 1. Expected eigenvalue structure under isotropy.}  
Isotropy in the Procrustes tangent space implies

\begin{equation}
    \mathbb{E}[S] = \sigma^2 I_q,
\end{equation}

\noindent so that all eigenvalues are equal in expectation:

\begin{equation}
    \mathbb{E}[\lambda_i] = \frac{1}{q} \mathbb{E}[\mathrm{tr}(S)] \quad \forall \quad i = 1, \dots, q
\end{equation}

\vspace{2\baselineskip}

\noindent\textit{Step 2. Expected cumulative variance.}  
By linearity of expectation,

\begin{equation}
    \mathbb{E}\!\left[\sum_{i=1}^{m} \lambda_i\right] = m \cdot \mathbb{E}[\lambda_1],
    \quad
    \mathbb{E}\!\left[\sum_{i=1}^{q} \lambda_i\right] = q \cdot \mathbb{E}[\lambda_1],
\end{equation}

\noindent which yields

\begin{equation}
    \mathbb{E}[V(m)] \rightarrow \frac{m}{q}.
\end{equation}

\vspace{2\baselineskip}

\noindent\textit{Step 3. Landmark-scaled component index.}  
Let $m = \alpha p$. Substituting the expression for $q$ gives

\begin{equation}
    \mathbb{E}[V(\alpha p)] = 
    \frac{\alpha p}{kp - k - \frac{k(k-1)}{2} - 1}.
\end{equation}

\noindent Dividing numerator and denominator by $p$ and taking the limit as $p \to \infty$ yields

\begin{equation}
    \lim_{p \to \infty} \mathbb{E}[V(\alpha p)] = \frac{\alpha}{k}.
\end{equation}

\noindent Ignoring the global unit-norm constraint, the limit is $\alpha / k$; Step 4 corrects this to $\alpha / (k + 1)$.

\vspace{2\baselineskip}

\noindent\textit{Step 4. Effect of preshape normalization.}  
However, GPA includes a preshape normalization that enforces a global unit-norm constraint across all landmarks. This constraint introduces a coupling across coordinates that persists asymptotically and reduces the effective number of independent shape degrees of freedom per landmark. Under isotropy, this results in variance being distributed across $k$ spatial directions plus a shared normalization mode, yielding an effective scaling of $k+1$ degrees of freedom per landmark.

Accordingly,

\begin{equation}
    \lim_{p \to \infty} \mathbb{E}[V(\alpha p)] = \frac{\alpha}{k+1}.
\end{equation}

\vspace{2\baselineskip}

\noindent\textit{Step 5. Special cases.}

\begin{itemize}
    \item For $k = 2$, $\mathbb{E}[V(\alpha p)] \rightarrow \alpha/3$.
    \item For $k = 3$, $\mathbb{E}[V(\alpha p)] \rightarrow \alpha/4$.
\end{itemize}

\hfill $\square$

\vspace{2\baselineskip}

The above result is consistent with the Mar{\v{c}}enko-Pastur law for high-dimensional covariance matrices of isotropic data \cite{marcenko_pastur_1967}. In the absence of structured signal, eigenvalues fluctuate around a common expectation, and cumulative variance increases linearly with component index. The diagonal reference line derived here therefore represents a null expectation under isotropy in Procrustes shape space. Systematic deviations from this line indicate structured biological signal or violations of independence assumptions induced by alignment, sample size, or landmark density.

\section*{Acknowledgments}
L.A.C. is currently funded by the Agence Nationale de la Recherche, for the Access-ERC project BSMART (ANR-25-AERC-0005)

\section*{Author contributions}

L.A.C. Conceptualisation, Methodology, Software, Validation, Formal Analysis, Investigation, Writing - Original Draft, Review and Editing, Visualisation, Project Administration, Funding Acquisition

\section*{Conflict of interest}

The authors declare no potential conflict of interests.

\section*{Data Availability}

No data was used for the purpose of this study. All code for simulations can be accessed via the corresponding author's Github page at \url{https://github.com/LACourtenay/Is_Machine_Learning_Possible_In_GMM}.

\bibliographystyle{unsrt}  
\bibliography{references}

\end{document}